# Maya a1: Do the Options Read the Model, or Does Their Position? Order Sensitivity in a Non-Autoregressive Decision Model

*Quantifying option-order dependence in Laya and removing most of it by shuffled fine-tuning*

Mohammed Nayeem Khan
mohammed.nayeem.khan@intel.com
mnknayeem@gmail.com

Logesh L
logesh.l@intel.com

Intel Corporation

**Abstract**

Laya answers typed multiple-choice questions in one forward pass by scoring a hidden state at one `[MASK]` marker per option. We re-engineer the checkpoint (421M parameters; a 2-layer option head with no positional tensor; a documented but unused `parallel` layout) and test whether the answer depends on where an option is listed. Presenting 951 Qwen-verified questions under every cyclic rotation of their options (or six random permutations for 12 options), the released checkpoint changes its answer for 25.8% [22.9–28.7] of questions (19.7% on the 605 English-language items alone); accuracy with the gold option first versus last differs by 10.1%, and the flips concentrate where the model's own confidence is low (54% versus 8% across confidence terciles). Fine-tuning with Laya's trainer on 2,818 teacher-verified rows with option shuffling cuts the flip rate to 7.2%; the identical rows without shuffling give 11.4%, so shuffling is responsible for 4.2% of the improvement. A confidence-weighted vote over orderings gives 69.7% against 69.1% for a single base pass at 4.1× the compute. Labels are LLM-made and the test sets are teacher-filtered. We also fine-tune with the authors' `parallel` layout on the same rows: flip rate 0.5%.

**Key findings (every number recomputed from frozen results)**

1. Base flip rate 25.8% [22.9–28.7]; Maya-v1 7.2%; same rows without shuffling 11.4%.
2. Gold-first vs gold-last accuracy gap 10.1% (base); flip rate by confidence tercile 54% / 16% / 8%.
3. Confidence-weighted vote (base): 69.7% vs 69.1% single pass, at 4.1× compute; single-pass Maya-v1: 86.3%.
4. Position dependence is acknowledged upstream (Laya docstring: slot logits of five identical options spread by 3.59). Authors' parallel layout, adapted on the same rows: flip rate 0.5%.

## 1 Introduction

A decision model that reads option texts should give the same answer when the same options are listed in a different order. Laya is a non-autoregressive encoder that answers typed questions in one forward pass; for a `choice` question it appends one `[MASK]` marker per option to the question and scores the hidden state at each marker. The options are therefore laid out along a single sequence, and nothing in the weights guarantees that an option's position is irrelevant.

We measure this on 951 Qwen-verified choice questions drawn from seven categories (3, 5 and 12 options), presenting each question under every cyclic rotation of its options (K≤5) or six random permutations (K=12), i.e. 3,893 model calls per checkpoint.

**Research questions.** RQ1: does the released checkpoint change its answer under reordering, and by how much? RQ2: what is the shape of the dependence (slot, option count, category, confidence)? RQ3: is a no-training remedy (voting over orderings) competitive? RQ4: does fine-tuning on Qwen-verified rows with option shuffling remove the dependence, and how much of the effect is due to shuffling itself rather than the extra data? RQ5: is the result stable across training seeds?

**Contributions.** (i) On the base checkpoint 25.8% [22.9–28.7] of questions change their answer purely because the options were reordered (19.7% on the 605 English-language items alone). (ii) The dependence is not a first/last artefact alone: the first-vs-last accuracy gap is 10.1%, but predictions are concentrated in particular slots (total-variation distance from uniform 0.09 at K=3, 0.07 at K=12). (iii) Fine-tuning with shuffling reduces the flip rate to 7.2%, whereas the same rows without shuffling give 11.4%. (iv) Confidence is an informative detector of flips (flip rate 54% / 16% / 8% across confidence terciles) and a confidence-weighted vote over orderings gives 69.7% versus 69.1% for a single base pass, at 4.1× compute. (v) We also fine-tune with the authors' `parallel` layout on the same rows: flip rate 0.5%.

## 2 Background and related work

**Position bias in generative models.** Multiple-choice evaluation of decoder LLMs is known to depend on where the answer is listed: models prefer particular option IDs or positions, and accuracy moves by several points when options are permuted [Zheng et al. 2023; Pezeshkpour and Hruschka 2023]; LLM judges prefer the first response shown [Wang et al. 2023]. Surface-form competition [Holtzman et al. 2021] adds that probability mass is spread over strings, not options. These analyses concern models that *generate* an option label; Laya does not generate, it reads one hidden state per option, so the mechanism, and the available remedies, differ.

**Order-invariant architectures.** Set-style encoders avoid order by construction. Laya's authors identified the issue in their own code: a docstring states that in the sequential layout "the slot logits of five identical options spread by 3.59" and the repository contains an unused `parallel` layout that shares position ids across options and masks cross-option attention. This paper does not replace that effort; it asks how large the effect is on realistic questions, what it looks like, and whether data alone (option shuffling during fine-tuning) removes most of it.

**Positional and surface-form effects beyond LLMs.** Option-order effects are not confined to decoder LLMs. In multiple-choice reading comprehension with fine-tuned encoders, answer position is a frequent artefact of dataset construction; here the dataset construction is uniform by design (every option is gold equally often in every slot), so any effect we measure belongs to the model. In LLM evaluation, the standard mitigation is to average over permutations at inference [Zheng et al. 2023; Wang et al. 2023]; we test that (Section 5.6) alongside a training-time mitigation and an architectural one.

**Why a separate study for a non-generative model.** The mitigations proposed for generative models, such as debiasing the prior over option IDs, assume a probability distribution over answer tokens. Laya has no such distribution: it has one scalar logit per slot. The relevant questions are therefore different: is the logit a function of the option text alone, or of its position? Is the dependence learned from the training order or built into the positional machinery of the encoder? The identical-option probe (Section 6) and the layout comparison (Section 7) answer these two questions separately.

**Evaluation protocol.** We follow the permutation-invariance literature in using all cyclic rotations for small K (so that each option occupies each slot exactly once) and random permutations for large K, and we report the flip rate (an item-level instability measure) next to mean accuracy, because accuracy averaged over orderings can hide that individual decisions change.

**Why a non-generative model could still be position-biased.** Encoders with position information (absolute or rotary) can learn position-dependent shortcuts whenever label frequency correlates with position during training. In Laya's own training recipe the options are presented in the order the question lists them unless the opt-in `--shuffle-options` flag is used (Laya's trainer docstring: "The scripts always present options in the order the question lists them, so the model can learn a position prior"). Our data-generation process makes gold labels uniform over slots by construction, so any position prior we measure is a property of the released weights, not of our data.

## 3 Re-engineering Laya

### 3.1 The artifact

Header inspection of the safetensors file enumerates 206 tensors and 421,293,830 parameters (842,609,210 bytes). 93.7% of the parameters are the encoder; the decision machinery on top of it (head, scorer, act head, type embeddings) is only 6.3%.

**Table RE-1.** Tensor groups of `convaiinnovations/laya/model.safetensors` as enumerated from the file header (all F16 except `temperature`, F32).

| Group | Tensors | Parameters | Share | Role (inferred from names and the open-source runtime) |
|---|---|---|---|---|
| encoder | 170 | 394,781,696 | 93.7% | ModernBERT-large: 28 layers, hidden 1024, vocabulary 50,368 |
| head | 24 | 25,192,448 | 6.0% | 2 post-encoder transformer layers (self-attention + 4096-wide FFN) over the option-slot states |
| scorer | 6 | 1,052,673 | 0.2% | LayerNorm → linear → scalar: one logit per option slot |
| act_head | 4 | 263,938 | 0.1% | small MLP (1028→256→2) that produces the 'act' probability |
| type_emb | 1 | 3,072 | 0.0% | one embedding each for choice / score / noul, added to the slot states |
| temperature | 1 | 3 | <0.01% | three F32 scalars, one per question type (all 1.0 in the base file; the live values are in the JSON config) |

### 3.2 Configuration

**Table RE-2.** Configuration keys of the base checkpoint that matter here (`rl_agent_config.json`).

| Key | Value | Consequence |
|---|---|---|
| encoder | answerdotai/ModernBERT-large | rotary positions; no learned absolute position table to inspect |
| max_len | 512 | total tokens: head + state |
| head_max_len | 192 | tokens available to instruction + all options |
| head_layers | 2 | slot states exchange information through self-attention |
| option_layout | sequential (default) | all published checkpoints use the sequential layout |
| temperature | 1.637, 1.251, 1.983 | per-type fallbacks (choice, score, noul) |
| temperature_by_options[choice:3-5] | 1.7602 | applied after clamping to [0.5, 5] |
| temperature_by_options[choice:6-10] | 1.0000 | applied after clamping to [0.5, 5] |
| temperature_by_options[score:3-5] | 1.2514 | applied after clamping to [0.5, 5] |
| temperature_by_options[noul:2] | 1.9834 | applied after clamping to [0.5, 5] |
| temperature_by_options[choice:11+] | 0.1006 | applied after clamping to [0.5, 5] — **clamped from 0.1006 to 0.5; still a 2× sharpener** |
| temperature_by_options[choice:2] | 1.9064 | applied after clamping to [0.5, 5] |

### 3.3 Runtime path

We read the open-source runtime (`laya` 0.4.26) alongside the weights; the path from request to probability is:

1. `Agent.predict_batch` (`agent.py` ~L1486) groups states that share a question set and tokenizes each state once.

2. `build_head` / `build_sequence` (`common.py` L157–L270) render `[CLS] <type> question: <instructions> [SEP] [MASK]` `opt`$_0$ … `[MASK] opt`$_{K-1}$ `[SEP] <state> [SEP]`; each option text is capped at 48 tokens, the head at `head_max_len`, and the state is cut from the *right* to fit `max_len` (`truncate_left=False`).
3. The encoder runs once. The hidden state at each option's `[MASK]` marker is taken as that option's slot state; the type embedding is added and the 2-layer head lets the slots attend to each other.
4. The scorer emits one logit per slot. For `choice` the logits are divided by the bucket temperature (`temp_bucket(type, K)`, clamped to [0.5, 5]) and soft-maxed; `score` uses the same softmax and reports an expected level; `noul` reports the probability of the second slot.
5. `unpermute_probs` maps probabilities back if the caller supplied `option_order`; the result carries `answer_confidence` (the top probability) and a separate entropy-based `confidence` that is explicitly not calibrated.

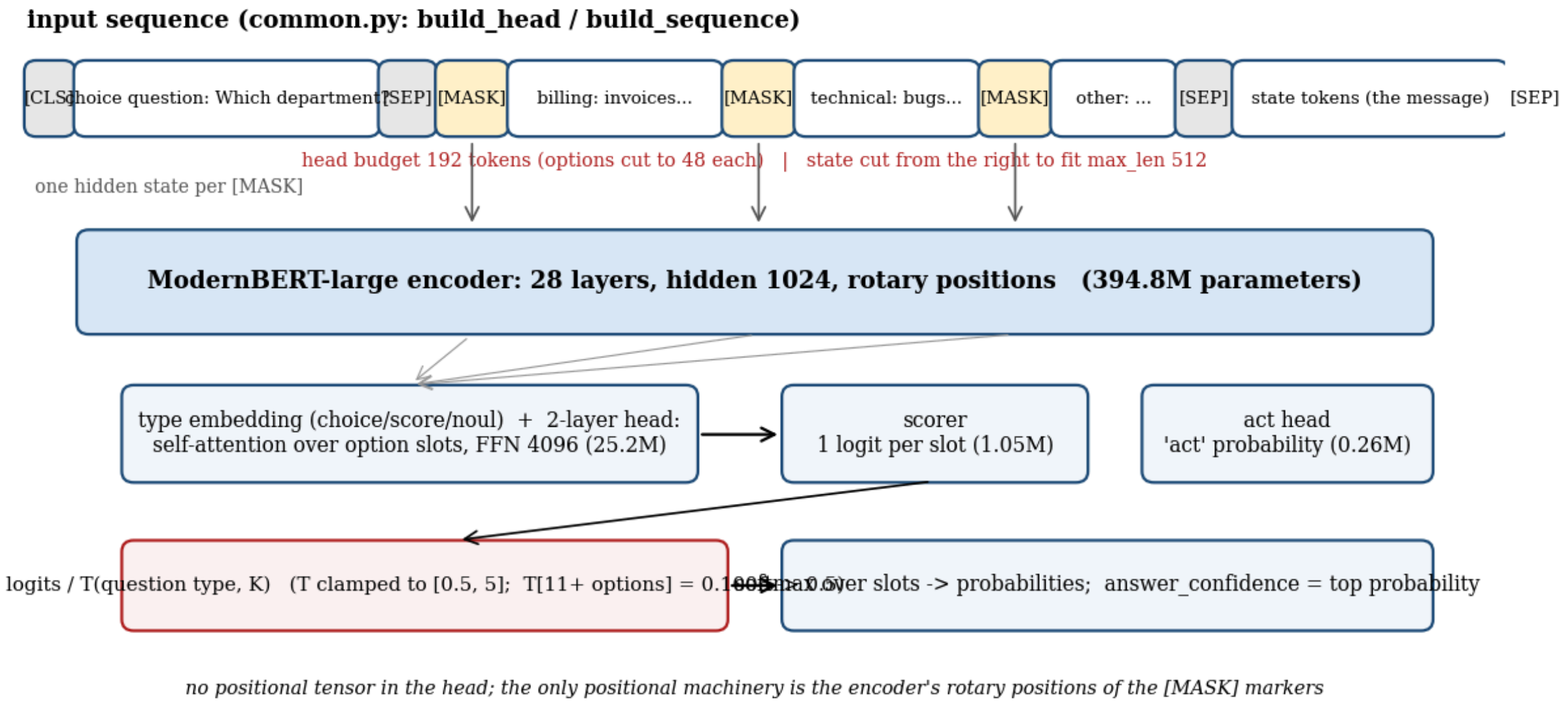

**Figure RE-1.** The decision path of Laya as reconstructed from the tensor inventory and the open-source runtime. Red: the two places where the studies in this series find problems (positional dependence through the encoder; the option-count temperature).

> **Note on the unused parallel layout.** `uses_parallel_layout(cfg)` reads `option_layout` from the config and defaults to `sequential`; the base config has no such key. The docstring in `common.py` says a checkpoint trained on one layout reads the other "as a different input", so switching layouts requires training, which is outside this paper.

**Provenance.** Header-inspection evidence `ev_2648a4101f7787d1303a…` (base, full-file SHA-256 `891102d372688fc2a094dac56a384bc537b87c63f21f9f3dac0be2b7cbc8d86c`) and `ev_fbf433f9d808bc103606…` (Maya-v1, `632868e784f96b313cb48cefa996dd78a392d27b056cf2a69d6798deb9ddab7c`). The tensor names, shapes and dtypes of the two files are **different**: all of our fixes change weight values and (for some) the JSON config, never the architecture. The inspection reports declared header facts only; it does not run the model, so everything behavioural below is measured black-box.

### 3.4 Hypotheses from the re-engineering

**H1 (position enters through the encoder).** The head has no positional-embedding tensor (Table RE-1 lists none): its two layers are permutation-equivariant over slots, and the type embedding is the same for every slot. The only way an option's *position* can influence its logit is through the encoder, whose rotary positions put each `[MASK]` marker at a different offset in the `sequential` layout. Prediction: answers change under reordering although the option texts are identical.

**H2 (the effect is largest for the first and last slot and for many options).** Rotary attention makes the first marker adjacent to the instruction and the last adjacent to the state; both are privileged. Prediction: a U-shaped accuracy curve over the gold slot, strongest at K=12.

**H3 (shuffled fine-tuning is sufficient).** If the dependence is learned rather than architectural, presenting options in random order during fine-tuning should remove it without touching the layout. Prediction: v1 flips far less than the no-shuffle ablation trained on identical rows.

## 4 Data, protocol and statistics

### 4.1 Held-out data from an LLM teacher

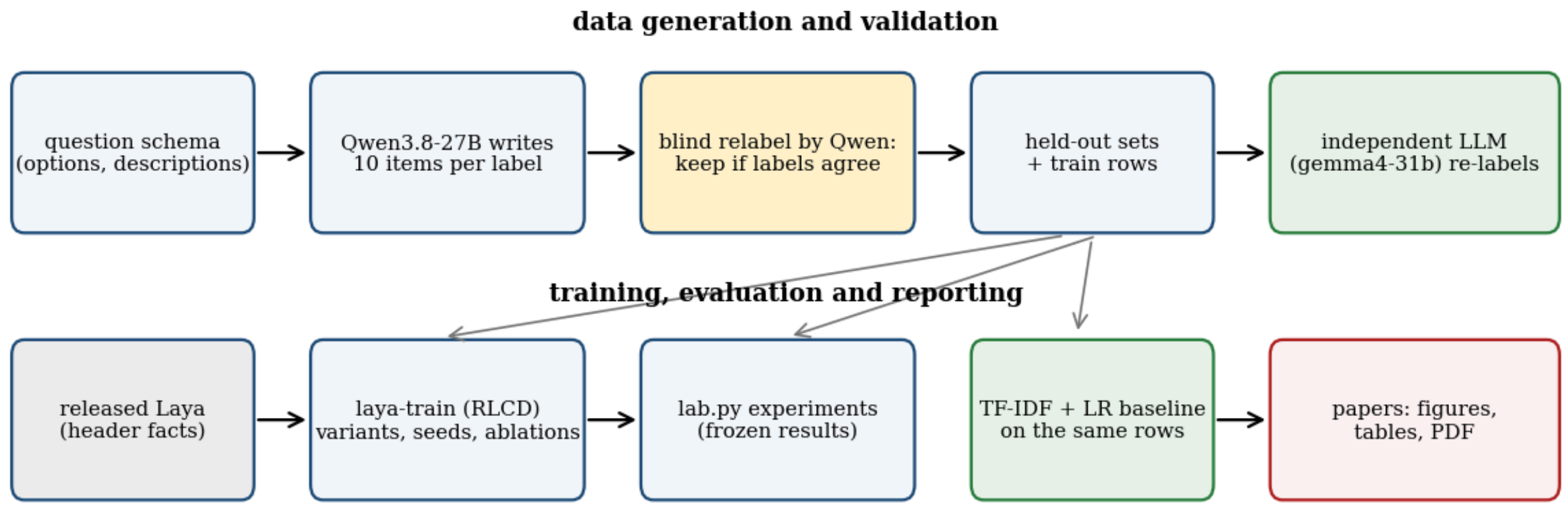

**Figure M-0.** Study design. Green boxes are the validity checks added after the first drafts: an independent LLM family re-labels every test item, and a classical baseline is trained on exactly the same rows.

We cannot ship human-labelled test sets for these gaps, so we generate them. For each category a prompt asks Qwen3.8-27B (thinking disabled) for ten realistic inputs *whose correct answer is a stated label*, with category hints (for example, "each text must contain a negation"). A second, independent call is shown the same inputs *without* the intended label and asked to choose; any input on which the two calls disagree is discarded. Exact prompts are in Appendix A. The filter removes inputs that are ambiguous to the teacher, but it also removes hard-but-valid cases, and it cannot remove errors the teacher makes consistently. **The test sets therefore measure agreement with Qwen, not human ground truth**; we return to this in each paper's threats section.

**Table M-1.** Candidates written by Qwen3.8-27B, and how many survive the blind re-labelling pass (generator-vs-relabeller agreement).

| Category | Generated | Kept | Agreement |
|---|---|---|---|
| hw_failure | 100 | 82 | 82.0% |
| json_state | 90 | 58 | 64.4% |
| many_hr | 773 | 716 | 92.6% |
| many_options | 240 | 195 | 81.2% |
| mix_Arabizi | 60 | 53 | 88.3% |
| mix_Hinglish | 60 | 58 | 96.7% |
| mix_Spanglish | 60 | 47 | 78.3% |
| sentiment_clean | 111 | 106 | 95.5% |
| urgency | 477 | 229 | 48.0% |
| ml_* (12 languages) | 1,420 | 1,289 | 90.8% |

**Table M-2.** Data splits (rows after de-duplication).

| Split | Rows | Use |
|---|---|---|
| train | 2,818 | fine-tuning (all Maya variants) |
| probe | 224 | first gap scan; reported before/after by laya-train |

| Split | Rows | Use |
|---|---|---|
| sent_test | 106 | held-out: sentiment, injection, long-context |
| json_test | 58 | held-out: JSON tickets |
| hw_test | 82 | held-out: hardware-validation failure class |
| many_test | 195 | held-out: 12-topic question |
| ml_test | 322 | held-out: 12 languages |
| urg_test | 78 | held-out: urgency (score) |
| ml_train / urg_train | 967 / 151 | extra rows used only by the multilingual / urgency variants |

### 4.2 Training

Every Maya model is Laya fine-tuned with its own trainer and loss (RLCD: a policy-gradient-style term over noisy logit samples rewarded by a proper scoring rule, plus soft cross-entropy), so the only experimental variables are the training rows and (where stated) the seed. Training takes minutes on one consumer GPU.

**Table M-3.** Fine-tuning recipe (`laya-train`; identical for every variant unless stated).

| Setting | Value |
|---|---|
| epochs | 4 |
| micro_batch | 8 |
| grad_accum | 8 |
| encoder_lr | 2.5e-05 |
| head_lr | 0.0001 |
| min_lr | 1e-06 |
| weight_decay | 0.01 |
| grad_clip | 1.0 |
| loss | rlcd |
| rl_samples | 4 |
| sigma_start | 0.4 |
| sigma_end | 0.1 |
| w_sph | 0.75 |
| w_rps | 1.0 |
| option shuffling | on for choice questions (off only in the no-shuffle ablation) |
| calibration holdout | carved from the training data by laya-train (fits temperatures; never the test sets) |
| hardware / software | 1× RTX 4090 (24 GB), torch 2.11.0+cu128, transformers 5.20.2, laya 0.4.26 |
| seed | 0 unless a seed is named |

### 4.3 Statistics

Proportions carry Wilson 95% intervals [Wilson 1927]. Two models on the same items are compared with the exact two-sided McNemar test [McNemar 1947]. Rates that are not per-item correctness (flip rate, ECE) use a row-level bootstrap with 1,000 resamples (300 where a statistic is expensive) [Efron and Tibshirani 1993]. We use $\alpha$=0.01 and a minimum practical effect of 5 points, fixed *before* the fix results were computed (see `papers.py`); no multiplicity correction is applied across the ten gaps we scanned, which is why we require both a small p-value and a non-trivial effect. ECE uses 10 equal-width bins unless stated.

### 4.4 Leakage audit

Exact duplicates between test and training rows are removed during generation. To check for near-duplicates we computed, for every test item, the character-5-gram Jaccard similarity to its most similar training row. Median maximum similarity is 0.08–0.15 across sets, and 0.1% of items on average reach 0.5 (worst set: 0.5%). Accuracy on the low-similarity subset equals overall accuracy to within 0.0 points, so we find no sign that gains come from memorised neighbours.

### 4.5 Order-sensitivity protocol

Questions are taken from every test set whose question type is `choice` with K≥3 (951 items; categories in Table 2). For K≤5 the options are rotated cyclically, giving K orderings in which every option appears in every slot once; for K=12 six random permutations are drawn once (seed 1) and reused for every item and every model, so all models see identical presentations. *Flip rate* is the fraction of items whose predicted option differs across orderings. *Gold-slot accuracy* averages correctness over all (item, ordering) pairs in which the gold option sits in a given slot. *Vote accuracy* counts an item as correct only if the gold option is the unique plurality answer over orderings. Flip rate has a non-zero floor on ambiguous items, so we always compare models on identical items and use the no-shuffle ablation as the matched control.

## 5 Results: how order-sensitive is Laya?

### 5.1 Headline

**Table 1.** Order sensitivity over 951 choice questions (K≥3), and over the 605 English-language items only (the 12 native-language categories are excluded in the third column because the English root is close to guessing on them). *Flip rate*: answer changes under at least one reordering. *Mean acc.*: accuracy averaged over all orderings. *Vote acc.*: confidence-weighted vote over all orderings (costs K× compute).

| Model | Flip rate, all [95% CI] | Flip rate, English only | Mean acc. | Vote acc. |
|---|---|---|---|---|
| Laya (root) | 25.8% [22.9–28.7] | 19.7% [16.5–22.8] | 69.1% | 69.7% |
| Laya typed-decisions | 12.5% [10.3–14.7] | 12.9% [10.2–15.5] | 71.0% | 71.8% |
| Laya-multilingual | 19.2% [16.9–21.7] | 26.0% [22.5–29.4] | 67.4% | 67.7% |
| Maya-v1 w/o option shuffle | 11.4% [9.5–13.6] | 8.3% [6.1–10.6] | 86.6% | 87.2% |
| Maya-v1 | 7.2% [5.7–8.9] | 4.5% [3.0–6.1] | 86.3% | 86.2% |
| Maya + parallel layout (authors' architectural fix) | 0.5% [0.1–1.1] | 0.3% [0.0–0.8] | 86.0% | 85.9% |
| Maya + name augmentation | 8.2% [6.5–9.8] | 6.0% [4.1–7.8] | 86.8% | 87.5% |

The root checkpoint flips on 25.8% [22.9–28.7] of all questions (gap criterion — lower bound >10%: **met**). The native-language categories (36% of items) are much less stable (36.4% flip) because the English root is close to guessing on them; excluding them, the root flips on 19.7% [16.5–22.8] of 605 English items, so the gap holds on English text alone. The authors' typed-decisions fine-tune flips on 12.5% (English only: 12.9%), and the shipped multilingual checkpoint on 19.2%, so the effect is not specific to one checkpoint.

### 5.2 By category and option count

**Table 2.** Per category: option count K, items, flip rate (base / no-shuffle ablation / Maya-v1) and mean accuracy (base / Maya-v1).

| Category | K | n | Flip base | Flip no-shuffle | Flip v1 | Acc base | Acc v1 |
|---|---|---|---|---|---|---|---|
| hw_failure | 5 | 115 | 33.0% | 10.4% | 4.3% | 61.9% | 96.9% |
| injection | 3 | 21 | 9.5% | 4.8% | 4.8% | 88.9% | 96.8% |
| json_state | 3 | 76 | 6.6% | 1.3% | 0.0% | 66.7% | 89.5% |
| many_options | 12 | 270 | 26.7% | 9.6% | 5.9% | 72.8% | 92.6% |
| ml_Arabic | 3 | 28 | 60.7% | 21.4% | 28.6% | 38.1% | 51.2% |

| Category | K | n | Flip base | Flip no-shuffle | Flip v1 | Acc base | Acc v1 |
|---|---|---|---|---|---|---|---|
| ml_Bengali | 3 | 26 | 84.6% | 15.4% | 19.2% | 24.4% | 37.2% |
| ml_Chinese | 3 | 29 | 20.7% | 6.9% | 10.3% | 77.0% | 95.4% |
| ml_German | 3 | 27 | 11.1% | 0.0% | 3.7% | 85.2% | 97.5% |
| ml_Hindi | 3 | 27 | 63.0% | 25.9% | 14.8% | 38.3% | 55.6% |
| ml_Japanese | 3 | 30 | 26.7% | 0.0% | 3.3% | 76.7% | 98.9% |
| ml_Korean | 3 | 21 | 38.1% | 33.3% | 23.8% | 52.4% | 60.3% |
| ml_Portuguese | 3 | 22 | 0.0% | 9.1% | 0.0% | 81.8% | 86.4% |
| ml_Russian | 3 | 28 | 32.1% | 0.0% | 3.6% | 67.9% | 95.2% |
| ml_Spanish | 3 | 28 | 0.0% | 14.3% | 14.3% | 71.4% | 81.0% |
| ml_Tamil | 3 | 28 | 89.3% | 64.3% | 10.7% | 35.7% | 39.3% |
| ml_Turkish | 3 | 28 | 17.9% | 10.7% | 7.1% | 66.7% | 86.9% |
| multilingual | 3 | 24 | 25.0% | 20.8% | 16.7% | 52.8% | 65.3% |
| negation | 3 | 17 | 0.0% | 17.6% | 11.8% | 88.2% | 88.2% |
| sentiment_clean | 3 | 106 | 1.9% | 6.6% | 2.8% | 94.0% | 95.6% |

Order sensitivity is far from uniform. The most unstable base categories are `ml_Tamil` (89%, n=28), `ml_Bengali` (85%, n=26), `ml_Hindi` (63%, n=27); the most stable are `ml_Portuguese` (0%, n=22), `ml_Spanish` (0%, n=28), `negation` (0%, n=17). Instability tracks difficulty: categories on which the base model is inaccurate are the ones that flip. Because the sets differ in difficulty, flip rate and accuracy are confounded, which is why every comparison in the fix section is made on identical items.

### 5.3 Where the gold option sits

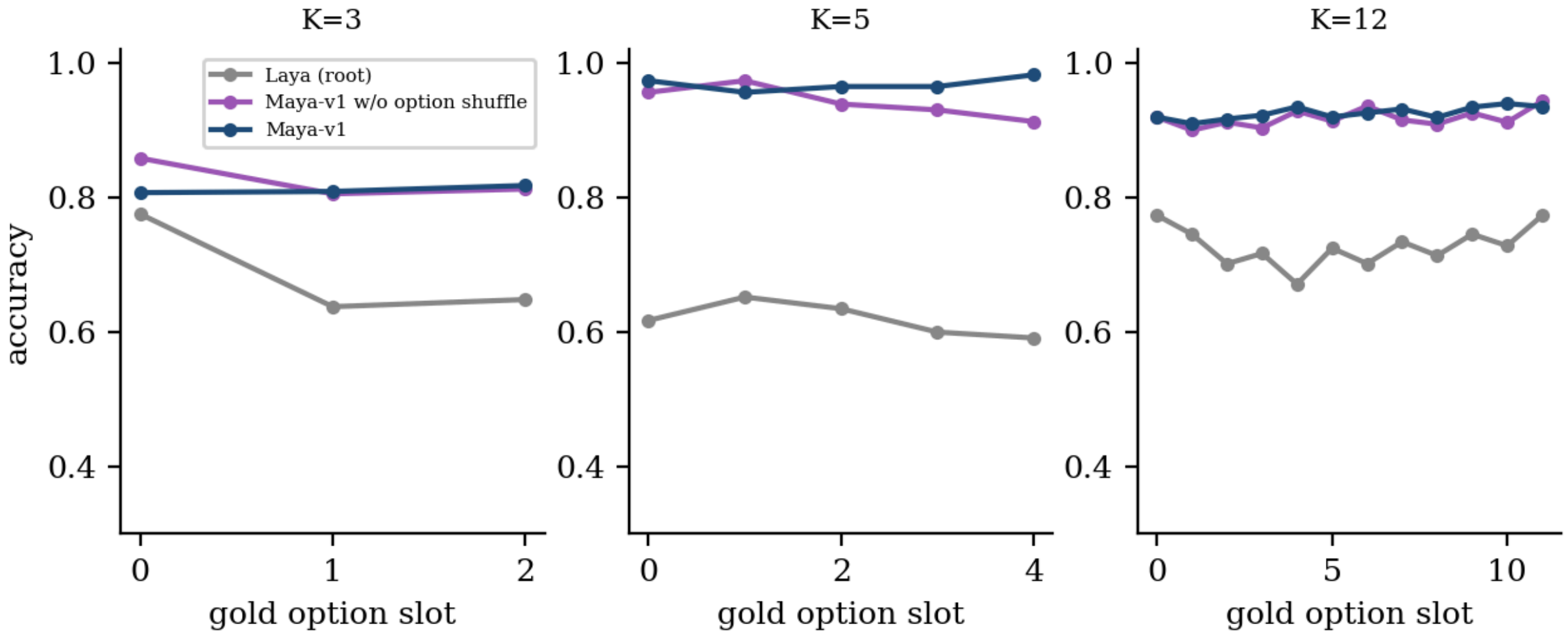

**Figure 1.** Accuracy as a function of the slot the gold option occupies, for the three option counts in the test pool.

Averaged over all items the base model is 77.1% accurate when the gold option is first and 67.0% when it is last (difference 10.1%). For K=12 the accuracy over the twelve gold slots ranges from 67% to 77% for the base model, 90% to 94% for fine-tuning without shuffling, and 91% to 94% with shuffling (Figure 1; with six permutations per item each slot has far fewer observations than at K=3, so individual points are noisy). First-vs-last accuracy without shuffling is 88.6% / 85.7% and with shuffling 85.3% / 86.9%.

### 5.4 Where the predictions land

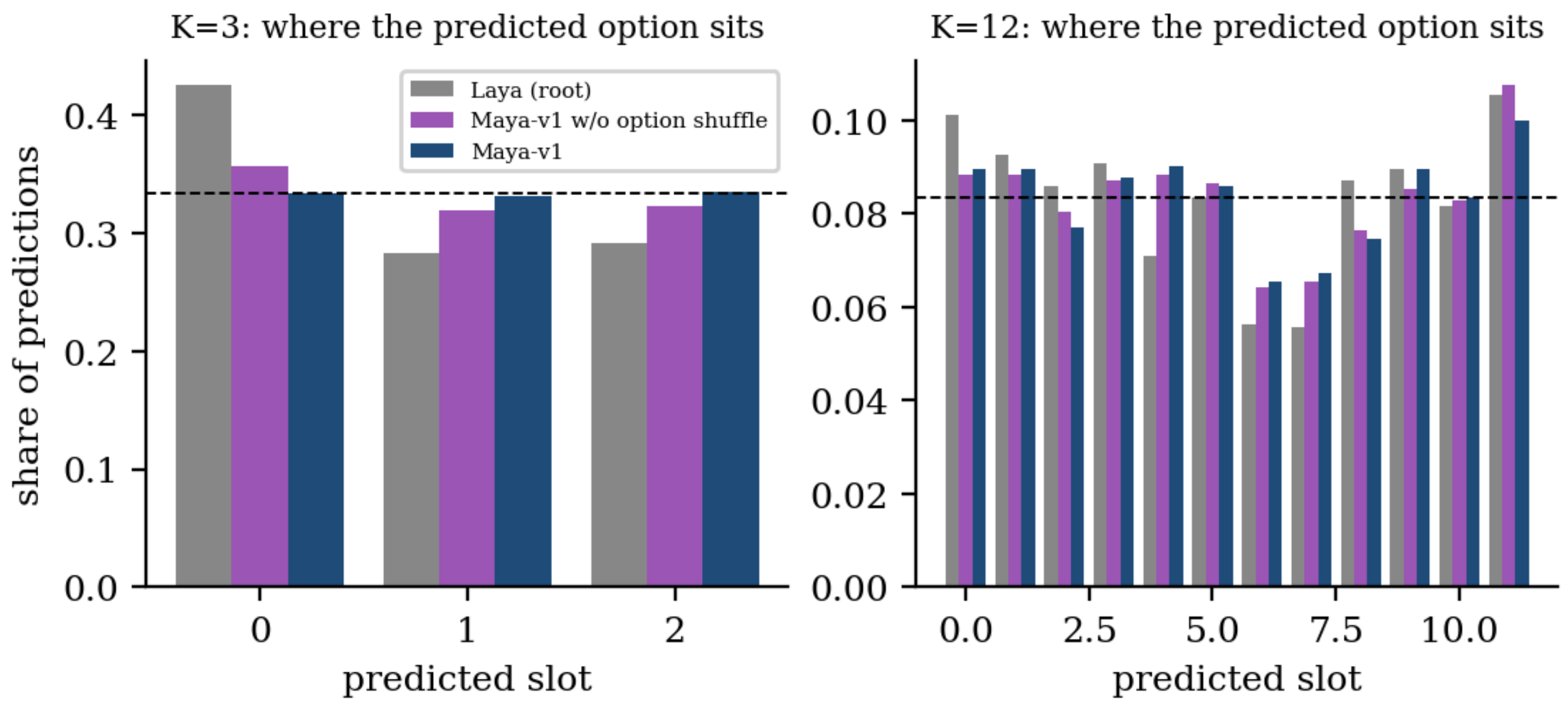

**Figure 2.** Distribution of the slot chosen by the model. Because every option appears in every slot equally often (rotations), an order-blind model would follow the dashed line apart from genuine errors.

Under rotations every option occupies every slot equally often, so the share of predictions per slot should be near 1/K for an order-blind model that is mostly right. The total-variation distance from uniform is 0.092 (K=3) and 0.069 (K=12) for the base model, 0.024 / 0.048 without shuffling and 0.002 / 0.049 with it.

### 5.5 Is the instability visible in the model's own confidence?

**Table 3.** Flip rate by tercile of the mean confidence the model reports across the orderings of that question (terciles fixed on the base model).

| Model | Lowest confidence | Middle | Highest |
|---|---|---|---|
| Laya (root) | 53.9% (n=317) | 15.8% (n=317) | 7.6% (n=317) |
| Maya-v1 | 73.7% (n=19) | 28.7% (n=171) | 0.7% (n=761) |

Largely yes. For the base model the flip rate falls from 54% in the least-confident tercile to 16% and 8%; among the 20% of questions on which it is most confident, 2.6% still flip. Confidence is therefore a useful, imperfect detector: a caller that abstains or re-queries with a shuffled order when confidence is low removes most unstable items, but 8% of even the most confident tercile can change answer under reordering. For Maya-v1 the corresponding figures are 74% / 29% / 1%, with most items in the top tercile.

### 5.6 A no-training alternative: voting over orderings

Because Laya exposes `option_order`, a caller can average over reorderings at inference time. Taking a confidence-weighted vote over all 4.1 orderings (on average) moves the base model from 69.1% (mean single-pass accuracy) to 69.7% (a gain of +0.6 points) at 4.1× the compute. A single pass of Maya-v1 reaches 86.3%; voting over Maya-v1 reaches 86.2%. Voting does not rescue the base model: its errors are largely consistent across orderings (the item is simply hard or out of its language), so averaging orderings cannot fix them; what shuffling during fine-tuning changes is the model itself.

### 5.7 How many orderings does an audit need?

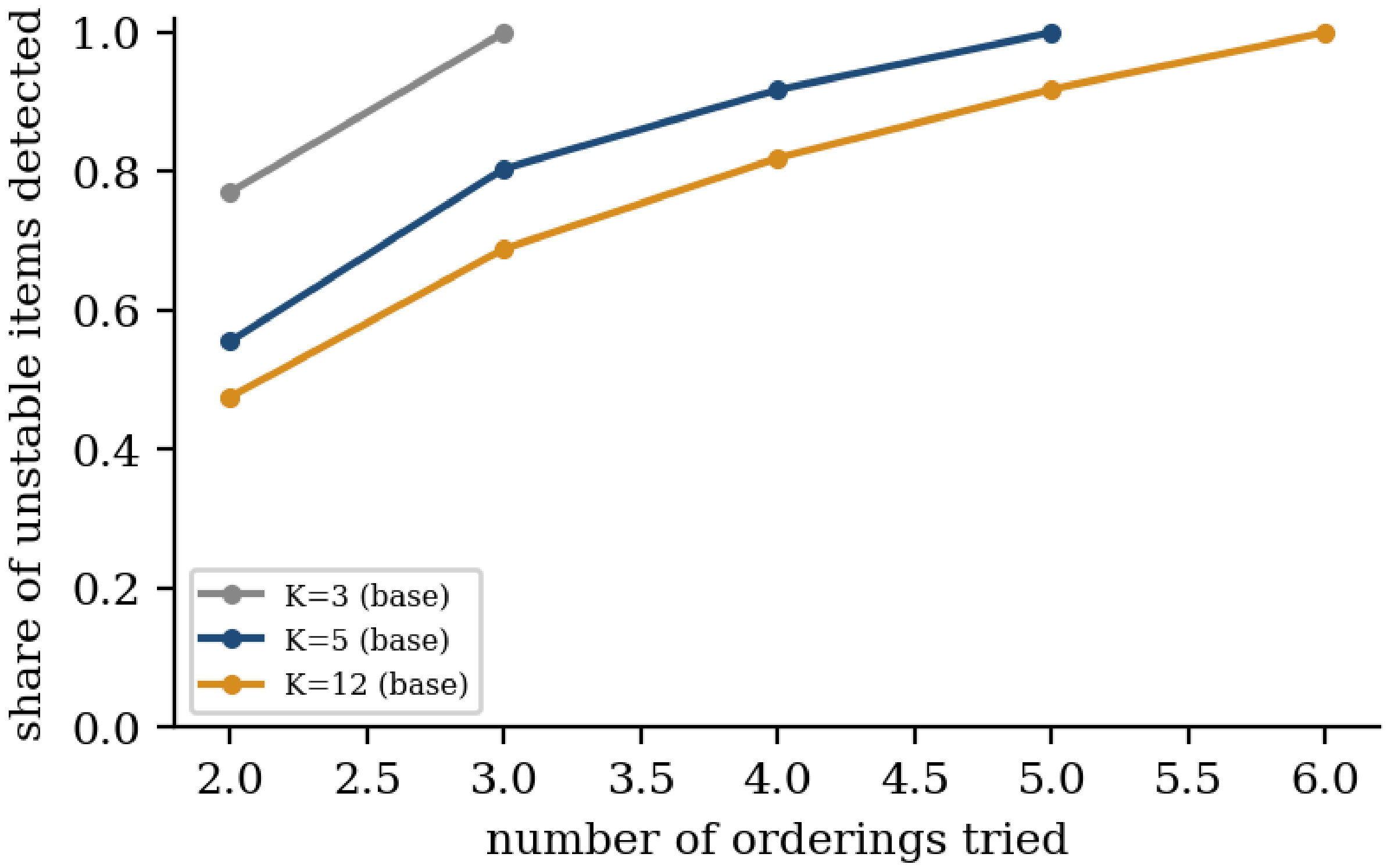

**Figure 3.** How many random orderings must be tried before an unstable item shows two different answers, by option count (base model; items that flip under the full set of orderings). K=3 has only three orderings, K=5 five, K=12 six.

A practical question for anyone who wants to *measure* order sensitivity on their own schema: trying two random orderings exposes 77% of the unstable items among the 3-option questions (the full set has only three orderings), 56% among 5-option and 48% among 12-option questions; four orderings expose 92% (K=5) and 82% (K=12). A two-ordering audit therefore misses a substantial share of unstable items on every option count; the figures above show how that share shrinks as more orderings are tried.

### 5.8 Unstable items are the wrong ones

Order sensitivity is not a cosmetic problem. The base model's mean accuracy is 41.6% on the 245 items that flip and 78.6% on those that do not; for Maya-v1 the figures are 43.7% on 68 flipping items and 89.6% on stable ones. Reducing flips therefore removes a pool of low-accuracy decisions rather than only reshuffling equally accurate ones.

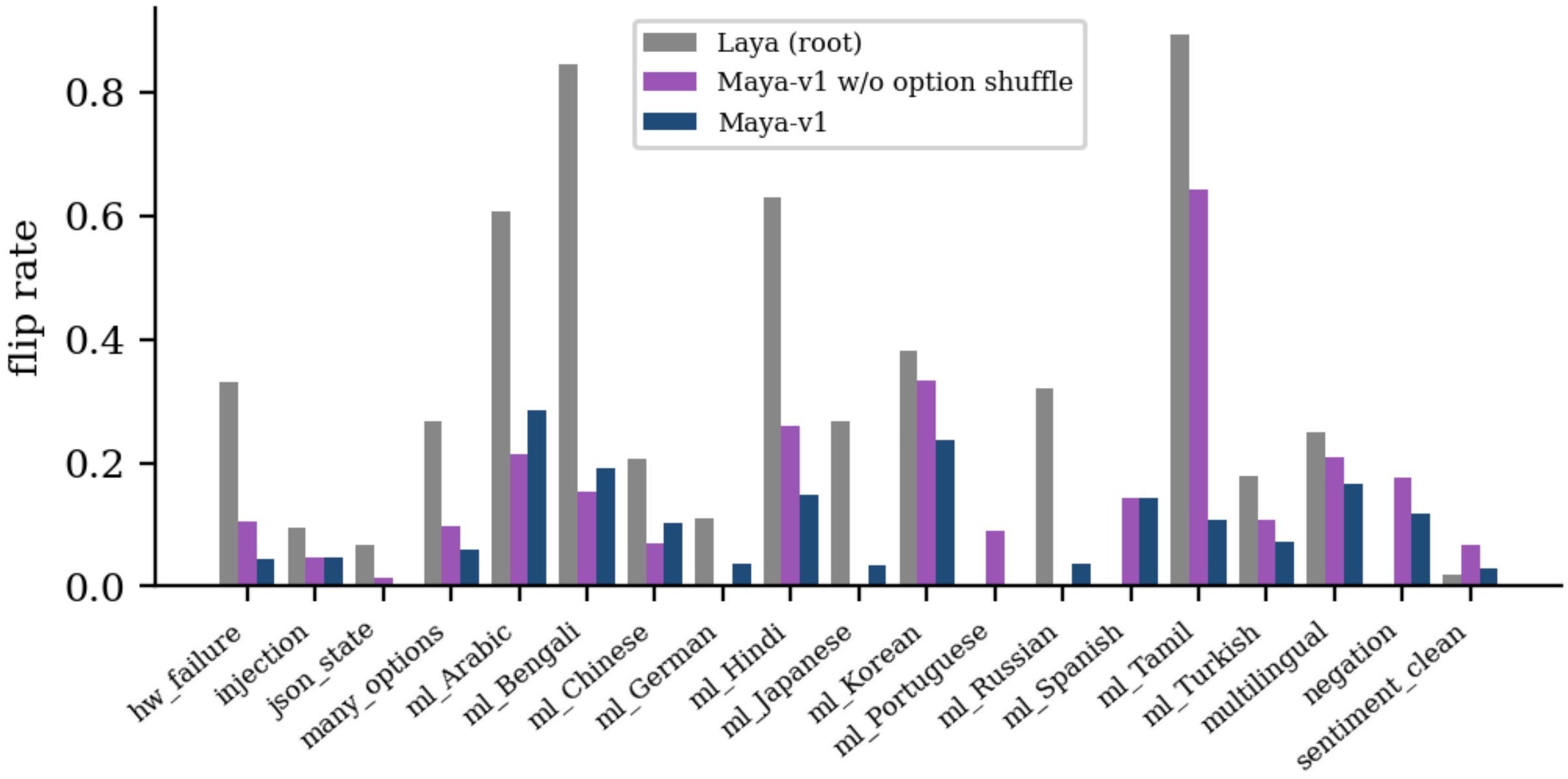

**Figure 4.** Flip rate by category.

## 6 Isolating the position prior

The flip rates measured so far mix two things: a content-independent position prior and interactions between position and content. A direct way to separate them is to give the model K options with *identical* descriptions, so that the only information about an option is its slot (and its name, which we randomise over slots).

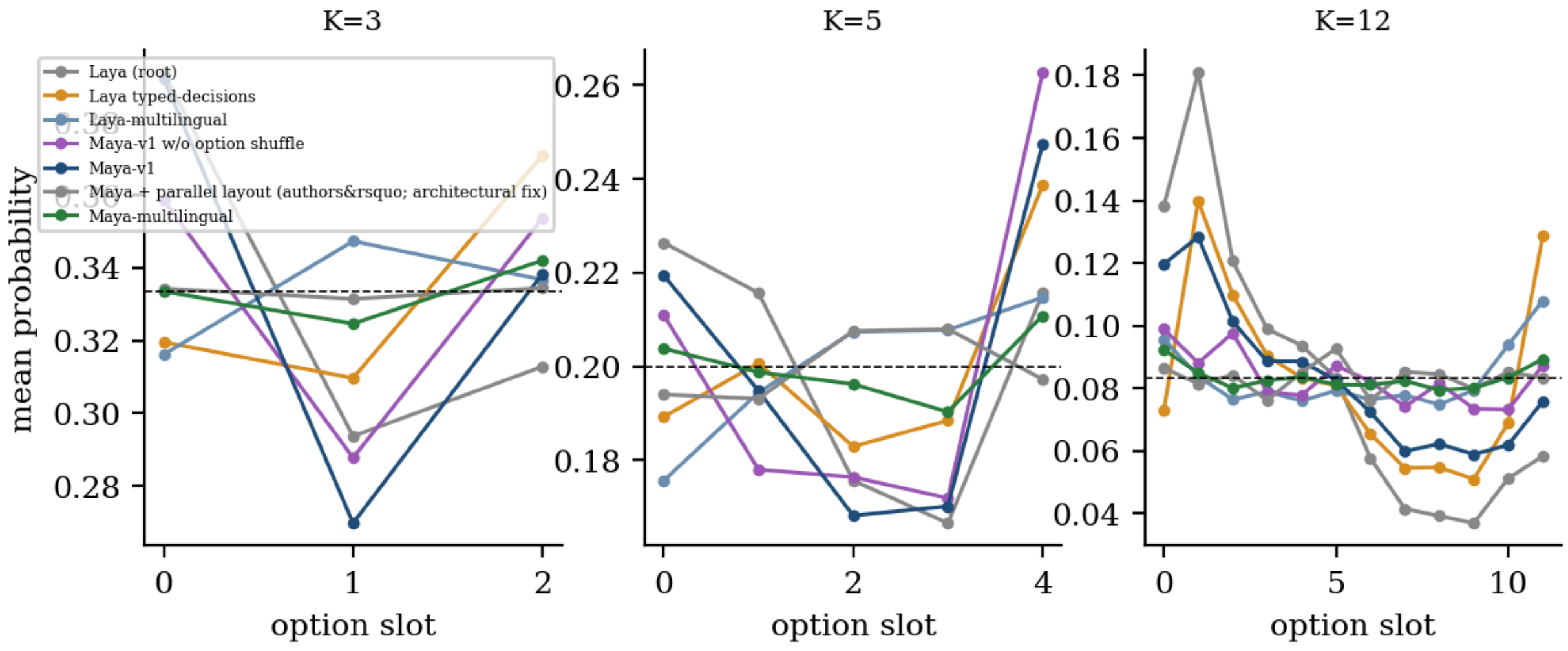

**Figure X-1.** The pure position prior: probability the model gives each slot when all K options have the *identical* description (letters are assigned to slots at random, so letter identity is balanced across slots). A model with no position prior follows the dashed line.

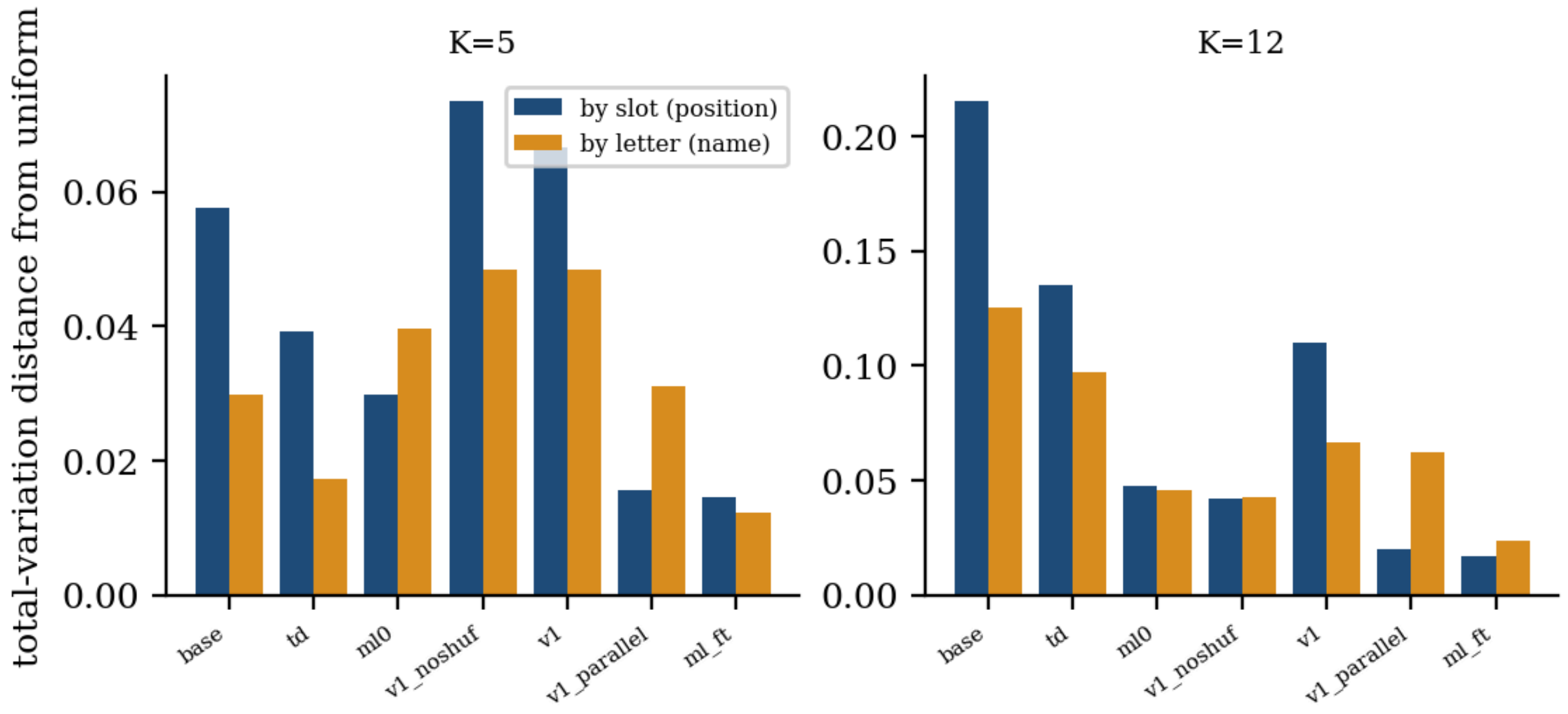

**Figure X-2.** Size of the position prior (blue) and of the option-name prior (orange), measured as total-variation distance of the mean probability from uniform.

**Table X-1.** Position prior (slot) and name prior (letter) with identical options, per option count K: total-variation distance from uniform, and the ratio of the most to the least probable slot.

| Model | K=3: slot TV | max/min | letter TV | K=5: slot TV | max/min | letter TV | K=12: slot TV | max/min | letter TV |
|---|---|---|---|---|---|---|---|---|---|
| Laya (root) | 0.060 | 1.3× | 0.064 | 0.058 | 1.4× | 0.030 | 0.216 | 4.9× | 0.125 |
| Laya typed-decisions | 0.038 | 1.2× | 0.032 | 0.039 | 1.3× | 0.017 | 0.135 | 2.7× | 0.097 |
| Laya-multilingual | 0.017 | 1.1× | 0.052 | 0.030 | 1.2× | 0.040 | 0.048 | 1.4× | 0.046 |
| Maya-v1 w/o option shuffle | 0.046 | 1.2× | 0.026 | 0.074 | 1.5× | 0.048 | 0.042 | 1.4× | 0.043 |
| Maya-v1 | 0.064 | 1.5× | 0.060 | 0.067 | 1.5× | 0.048 | 0.110 | 2.2× | 0.067 |
| Maya + parallel layout (authors' architectural fix) | 0.002 | 1.0× | 0.037 | 0.016 | 1.1× | 0.031 | 0.020 | 1.2× | 0.062 |
| Maya-multilingual | 0.009 | 1.1× | 0.023 | 0.015 | 1.1× | 0.012 | 0.017 | 1.2× | 0.024 |

With identical options the content carries no signal, so any preference is pure prior. The base checkpoint is far from uniform at K=12: its most probable slot receives 18% against 4% for the least probable (uniform: 8.3%), a 4.9× ratio. Fine-tuning with shuffling (Maya-v1) lowers the ratio only to 2.2× and the no-shuffle ablation to 1.4×; the parallel layout brings it to 1.22× (essentially flat). **This qualifies the headline result of the previous section:** shuffled fine-tuning removes most *content-dependent* flips on realistic questions, but a latent position prior remains whenever the content is uninformative, and only the architectural change removes it. The letter prior (orange in Figure X-2) is a separate effect: the model also prefers some option *names*, and the parallel layout does not remove that.

## 6.2 The 12-option question with 24 orderings

The six orderings used earlier give each slot few observations at K=12. Here we use 24 random orderings per item, which gives about 540 (item, ordering) observations per gold slot and allows a proper significance test of the slot effect.

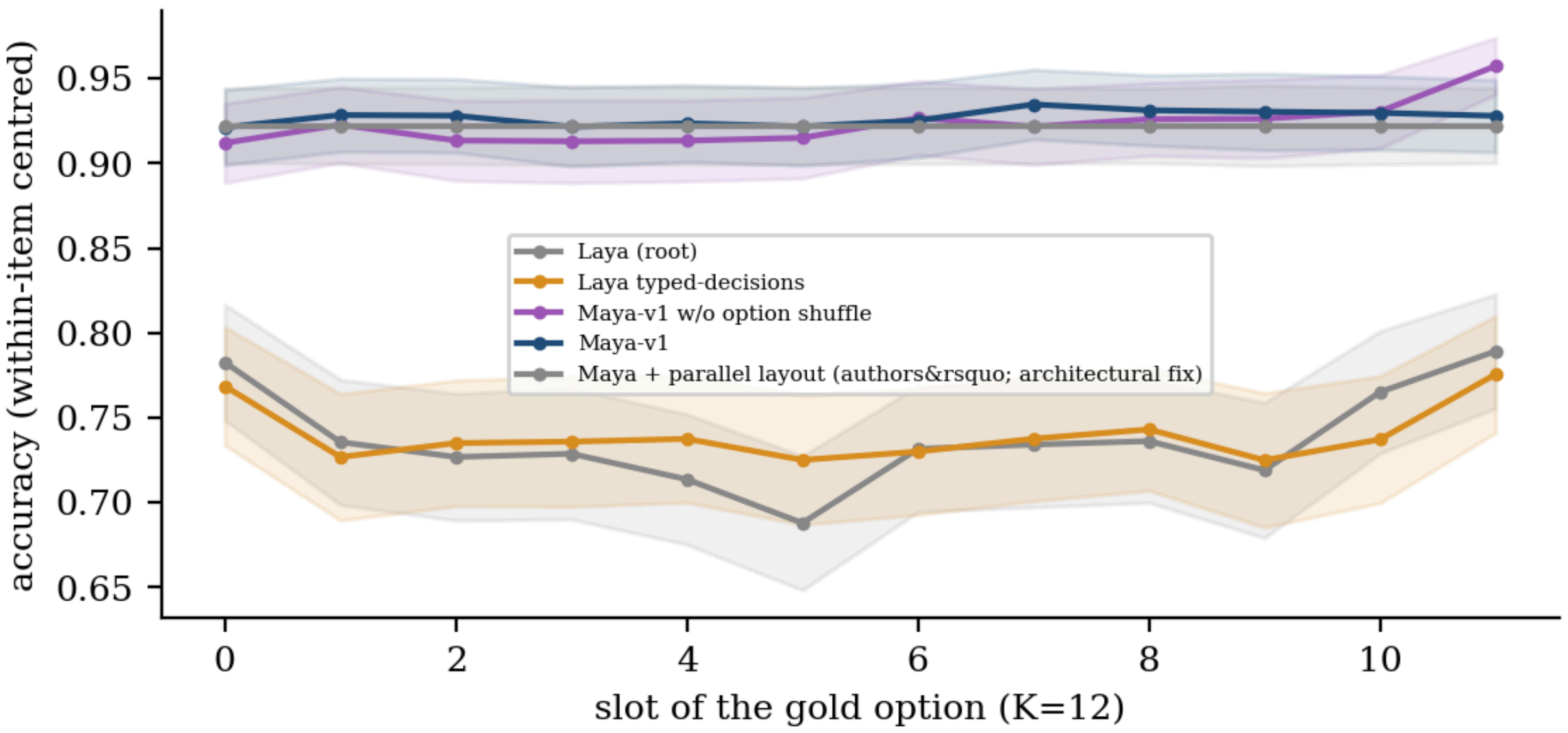

**Figure X-3.** Accuracy by the slot the gold option occupies on the 12-topic question, with 24 random orderings per item (≈540 observations per slot); bands are 95% normal intervals.

**Table X-2.** The 12-option question under 24 random orderings per item (270 items): flip rate, mean accuracy, spread of the within-item-centred slot accuracies (max−min) and the permutation test for any slot effect (variance of centred slot accuracies against 2000 within-item shuffles). Centring removes item difficulty, which would otherwise be confounded with slot because the same 24 permutations are used for every item; an exactly order-invariant model has zero centred spread.

| Model | Flip rate [95% CI] | Mean acc. | Centred slot spread | Slot-effect p |
|---|---|---|---|---|
| Laya (root) | 35.2% [29.6–41.1] | 73.8% | 10.1% | p<0.001 |
| Laya typed-decisions | 26.3% [21.1–32.2] | 74.0% | 5.1% | p<0.001 |
| Maya-v1 w/o option shuffle | 13.0% [9.3–17.2] | 92.3% | 4.5% | p<0.001 |
| Maya-v1 | 10.7% [7.0–14.6] | 92.7% | 1.4% | p=0.573 |
| Maya + parallel layout (authors' architectural fix) | 0.0% [0.0–0.0] | 92.2% | 0.0% | p=1.000 |

A slot effect on accuracy is statistically detectable (permutation p<0.01) for Laya (root), Laya typed-decisions, Maya-v1 w/o option shuffle. Any residual slot dependence of the parallel-layout model is not significant at this sample size.

### 6.3 An LLM on the same protocol

Position bias is usually reported for generative LLMs [Zheng et al. 2023]. To put Laya's numbers in context, we ran the same rotation protocol on an independent instruction-tuned LLM.

**Table X-3.** Order sensitivity of an independent LLM (gemma4-31b, every cyclic rotation) next to Laya on the same items: flip rate and mean accuracy for the LLM, flip rate for Laya.

| Set | n | LLM flip rate | LLM mean acc. | Laya (root) flip | Maya-v1 flip | Maya + parallel layout (authors' architectural fix) flip |
|---|---|---|---|---|---|---|
| sentiment (K=3) | 106 | 0.9% | 99.4% | 1.9% | 2.8% | 0.9% |
| hardware class (K=5) | 82 | 2.4% | 96.6% | 31.7% | 2.4% | 0.0% |
| JSON tickets (K=3) | 58 | 0.0% | 96.6% | 5.2% | 0.0% | 1.7% |

The comparison shows whether the order sensitivity of the encoder is unusual: where the LLM flips on a few percent of items and the base Laya on several times more, the encoder is clearly the less order-robust of the two; where they are similar, order sensitivity is a general property of option-list classifiers.

### 6.4 All categories and seeds at a glance

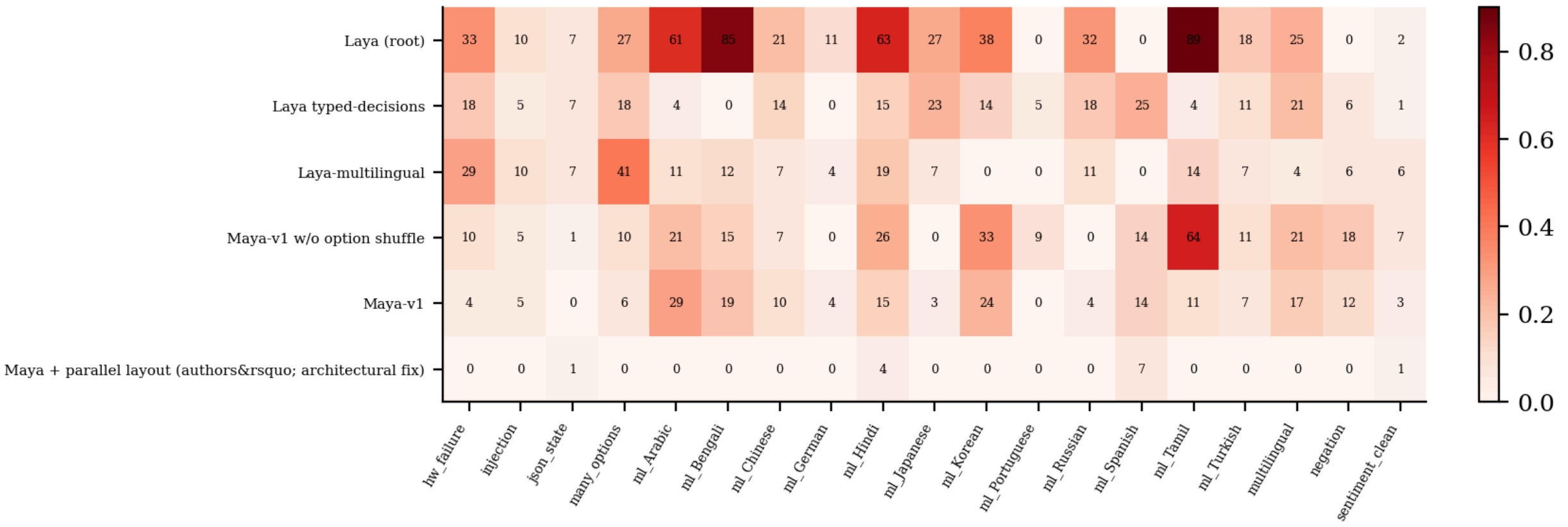

**Figure X-4.** Flip rate (%) by category (columns) and model (rows). Native-language categories dominate the base model's instability; the parallel-layout row is near zero everywhere.

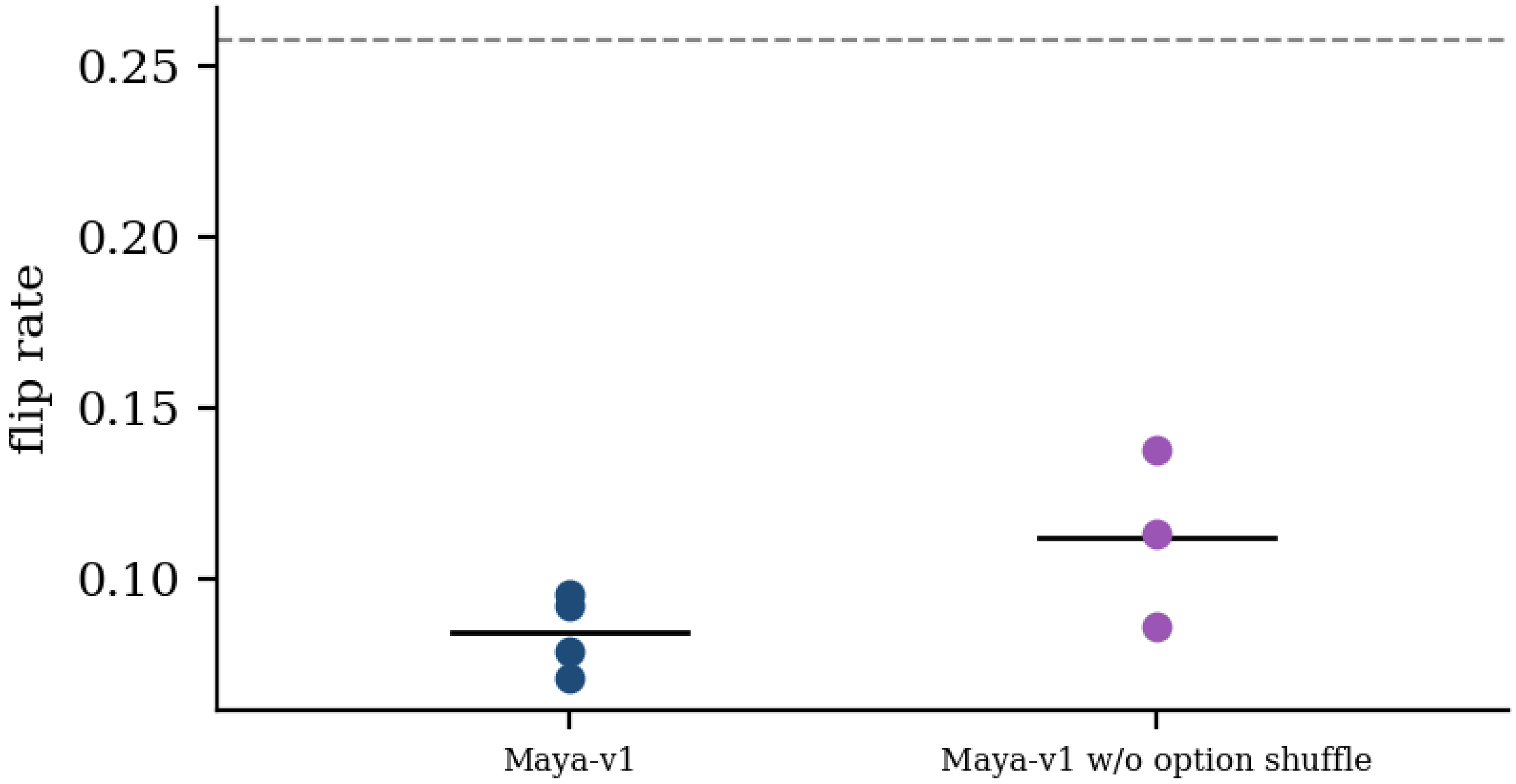

**Figure X-5.** Flip rate for every training seed (dots) and their mean (bar); the dashed line is the base model.

## 7 Fix: fine-tuning with option shuffling

Maya-v1 is Laya fine-tuned on 2,818 Qwen-verified rows (Table M-3) with the trainer's `--shuffle-options`, which re-encodes each choice item every epoch under a random option order and permutes the target accordingly. Its flip rate is 7.2% [5.7–8.9]. The paired reduction relative to the base model is 18.6% [15.6–21.7]; fix criterion (≥5 points, interval excluding 0): **met**.

### 7.1 Attribution: shuffling versus extra data

Training on exactly the same rows without shuffling gives a flip rate of 11.4% [9.5–13.6]. The difference attributable to shuffling is 4.2% [2.2–6.4]. The remainder of the improvement over the base (14.4%) is what fine-tuning on these rows alone buys, presumably because it teaches the model to rely on option content for these schemas. Attribution is therefore not all-or-nothing: shuffling matters, and so does task data.

### 7.2 Seed stability

**Table 4.** Flip rate across training seeds (seed 0 is the model reported elsewhere).

| Training recipe | Seeds | Flip rate per seed | Mean ± sd (pts) |
|---|---|---|---|

| Training recipe | Seeds | Flip rate per seed | Mean ± sd (pts) |
|---|---|---|---|
| Maya-v1 | 4 | 7.2%, 9.6%, 7.9%, 9.3% | 8.5% ± 1.1 |
| Maya-v1 w/o option shuffle | 3 | 11.4%, 13.8%, 8.6% | 11.3% ± 2.6 |

Seed-to-seed variation is small compared with the gap to the base model; the conclusion does not depend on the particular seed we chose first.

### 7.3 Cost

Accuracy averaged over orderings moves from 69.1% to 86.3%. The fix therefore does not trade accuracy for stability on these schemas. Training takes about five minutes on one consumer GPU; inference cost is unchanged (one pass).

### 7.4 The authors' architectural fix: the parallel layout

The trainer exposes `option_layout`; with `"parallel"` every option starts at the same position id and options may not attend to one another, so reordering only reorders marker embeddings. We fine-tuned the base checkpoint with exactly the Maya-v1 rows and recipe except for this layout (`train_parallel.py`; same seed, shuffling on). Its flip rate is 0.5% [0.1–1.1], versus 7.2% for the sequential Maya-v1 (paired difference sequential − parallel 6.6% [4.9–8.4]) and 25.8% for the base model ([22.4–28.2] reduction). Mean accuracy over orderings is 86.0% versus 86.3%. Its flip rate is essentially zero, as expected if the layout makes the encoder blind to option order; the data-side fix (shuffling) therefore recovers 74% of the architectural fix's reduction.

## 8 Hypothesis scorecard

Each hypothesis from Section 3, with the outcome the data support.

**Table H-1.** Scorecard of the hypotheses stated in the re-engineering section.

| # | Hypothesis | Outcome | Evidence |
|---|---|---|---|
| H1 | Position enters through the encoder, not the head (head has no positional tensor). | supported | Flips: base 25.8%; adapting the base to the parallel layout (which only changes encoder positions/masks) gives 0.5%. Identical-option prior at K=12: base 4.9×, parallel 1.22×. |
| H2 | A U-shaped accuracy curve over the gold slot, strongest at K=12. | partly supported | Within-item-centred slot effects are statistically detectable for the sequential checkpoints that were not shuffle-trained (Table X-2) and shrink or vanish with shuffling; the shape is not simply U-shaped (Figure X-3). |
| H3 | Shuffled fine-tuning is sufficient to remove the dependence. | partly supported | It removes most content-dependent flips (25.8% → 7.2%; no-shuffle 11.4%) but leaves a latent prior with identical options (2.2× at K=12 vs 4.9× for the base). |

## 9 Validity of the labels and comparison with an LLM

Every test label in this series was written by one LLM family (Qwen) and filtered by that same family. To check how teacher-specific the labels are we had a different model (gemma4-31b) label every held-out item blind, with the same prompt format as the filter.

**Table V-1.** Agreement of an independent LLM (gemma4-31b, a different model family from the Qwen teacher) with the Qwen-made gold labels, labelling every held-out item blind. High agreement means the labels are not specific to the teacher; the remainder is the part of the label noise that two strong models disagree on.

| Set | n | Agreement [95% CI] |
|---|---|---|
| probe (7 categories) | 224 | 97.3% [94.3–98.8] |
| sentiment | 106 | 99.1% [94.8–99.8] |
| hardware | 82 | 95.1% [88.1–98.1] |
| 12 topics | 195 | 94.4% [90.2–96.8] |
| JSON tickets | 58 | 96.6% [88.3–99.0] |

**Table V-2.** Independent LLM used as a *classifier* versus the released and the retrained Laya on identical items (accuracy against the Qwen gold; last column: exact McNemar, LLM vs retrained Laya).

| Set | gemma4-31b (LLM) | Laya (root) | Maya (fine-tuned) | LLM vs Maya |
|---|---|---|---|---|
| sentiment | 99.1% [94.8–99.8] | 93.4% [87.0–96.8] | 95.3% [89.4–98.0] | p=0.219 |
| hardware | 95.1% [88.1–98.1] | 62.2% [51.4–71.9] | 96.3% [89.8–98.7] | p=1.000 |
| 12 topics | 94.4% [90.2–96.8] | 73.3% [66.7–79.0] | 92.8% [88.3–95.7] | p=0.629 |
| JSON tickets | 96.6% [88.3–99.0] | 74.1% [61.6–83.7] | 89.7% [79.2–95.2] | p=0.219 |
| probe (7 categories) | 97.3% [94.3–98.8] | 71.9% [65.7–77.4] | 91.1% [86.6–94.1] | p=0.004 |

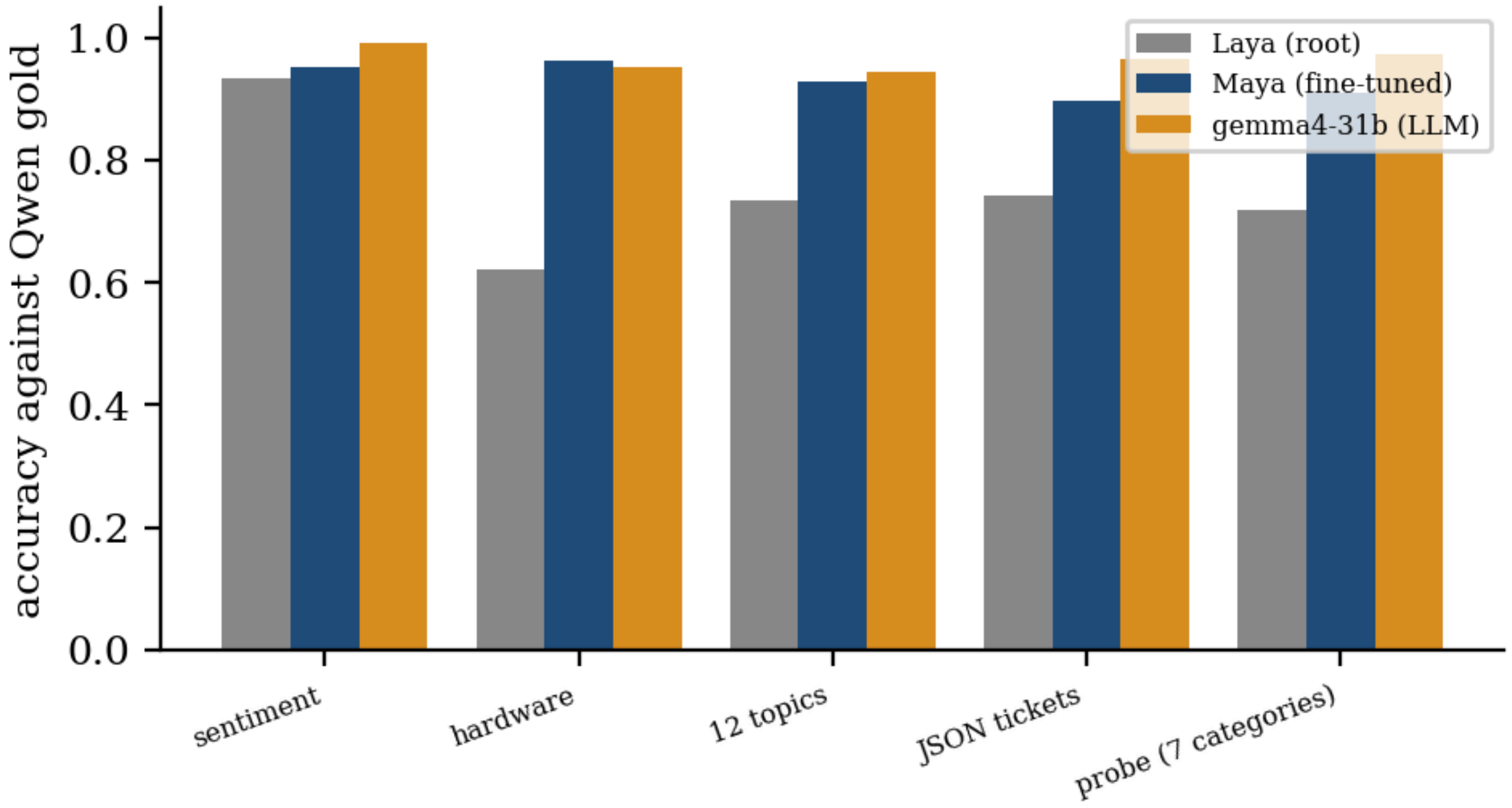

**Figure V-1.** Accuracy of the released Laya, the retrained Laya and an independent LLM on identical items.

**Cost.** One Laya forward pass costs about 2 ms per item on an RTX 4090 (batch 16); the LLM took a median 214 ms per item when ten items were sent in one request over the network (90× slower), and needs a hosted model. The endpoint latency is not a hardware-matched comparison, but the order of magnitude is the point.

## 10 Statistical power and sensitivity of the verdict

On 951 paired items, Maya-v1 (items that stay stable under reordering) is better than base by +18.6 points (bootstrap 95% interval +15.6 to +21.7; exact McNemar p<0.001); 215 items are fixed and 38 broken, a discordance rate of 26.6%.

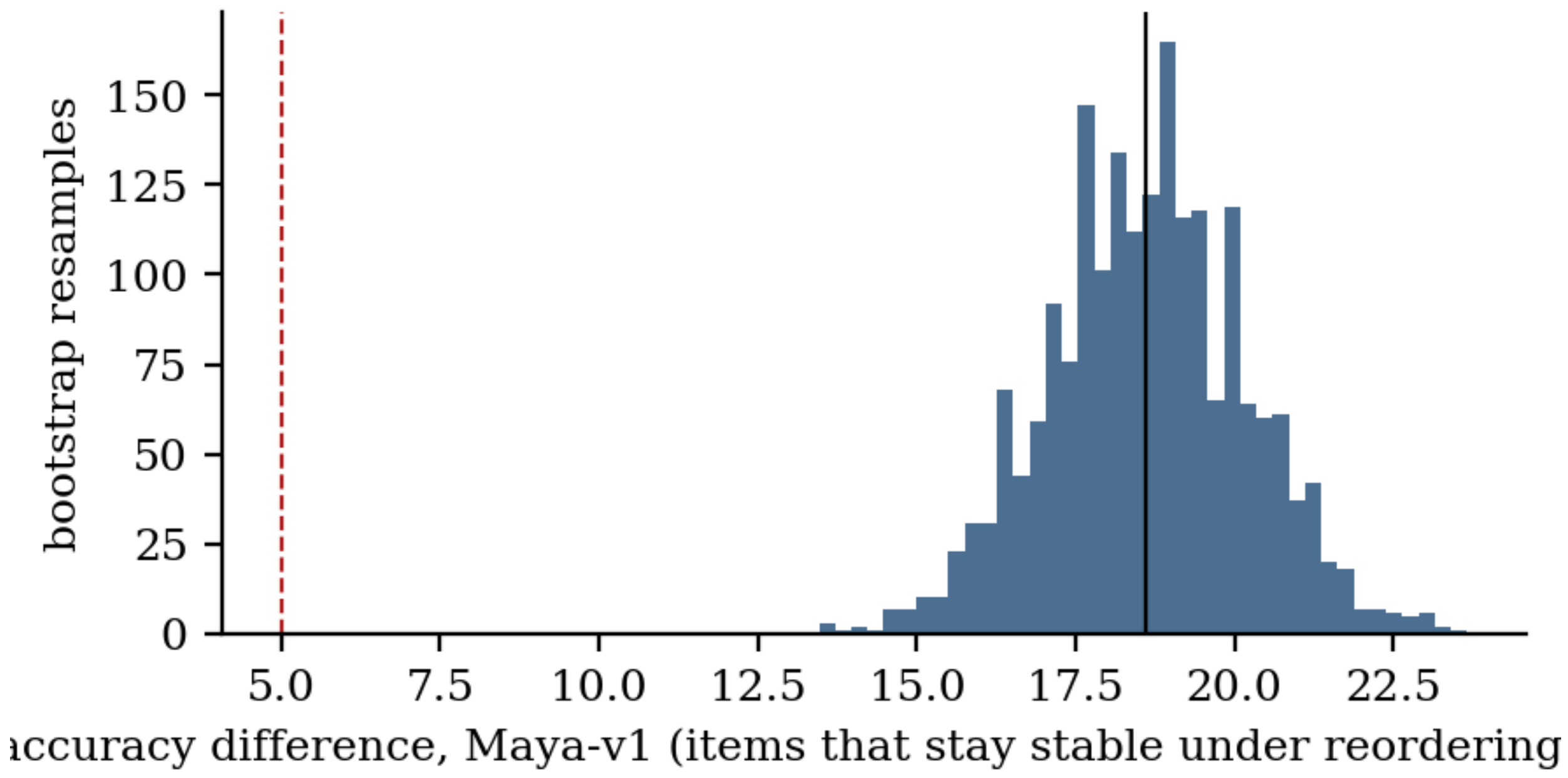

**Figure P-a1.** Bootstrap distribution (2000 resamples) of the paired difference; dashed line = the 5-point practical threshold.

**Table P-a1.** Sensitivity of the verdict for Maya-v1 (items that stay stable under reordering) vs base to the decision thresholds: fix is 'met' if McNemar p<α and the difference is at least the stated number of points. Last column: minimum detectable effect (points) at 80% power for this n and discordance rate.

| Significance | ≥2 pts | ≥5 pts | ≥10 pts | MDE (pts) |
|---|---|---|---|---|
| α=0.05 | met | met | met | 4.7 |
| α=0.01 | met | met | met | 5.7 |
| α=0.001 | met | met | met | 6.9 |

The result is robust to the choice of thresholds wherever Table P-a1 says 'met'; where it does not, the effect is smaller than the power of this test set can resolve.

## 11 Collateral effects and inference cost

A fix that helps one gap but breaks another is not a fix. Table C-1 reports every variant on every held-out set; the cost table shows what the change costs at inference.

**Table C-1.** Collateral check: accuracy on every held-out set. An asterisk marks a difference from the base checkpoint with exact McNemar p<0.01. Columns: probe (n=224), sent_test (n=106), json_test (n=58), hw_test (n=82), many_test (n=195), ml_test (n=322), urg_test (n=78).

| Model | probe | sent | json | hw | many | ml | urg |
|---|---|---|---|---|---|---|---|
| Laya (root) | 72% | 93% | 74% | 62% | 73% | 61% | 58% |
| Maya-v1 w/o option shuffle | 91% * | 93% | 100% * | 93% * | 92% * | 74% * | 87% * |
| Maya-v1 | 91% * | 95% | 90% | 96% * | 93% * | 74% * | 91% * |
| Maya + name augmentation | 91% * | 92% | 97% * | 89% * | 93% * | 79% * | 90% * |
| Maya + injection augmentation | 92% * | 96% | 100% * | 98% * | 95% * | 77% * | 83% * |
| Maya + long-context (1024) | 91% * | 96% | 97% * | 94% * | 92% * | 78% * | 92% * |

**Table K-1.** Inference cost on one RTX 4090 (batch 16, mixed short items, median of three passes). Fine-tuning does not change the architecture, so Maya models cost the same as the checkpoint they start from.

| Checkpoint | Items / s | ms / item | Peak GPU MB |
|---|---|---|---|
| Laya (root) | 630 | 1.6 | 2468 |

| Checkpoint | Items / s | ms / item | Peak GPU MB |
|---|---|---|---|
| Laya-multilingual | 831 | 1.2 | 1588 |
| Maya-v1 | 318 | 3.1 | 2465 |

## 12 Error analysis

The following items flip under the base model but not under Maya-v1 (first six orderings shown; `pred` is the option name the model returned).

> **Item 76 (multilingual, K=3, gold `technical`).** “ছবি আপলোড করার সময় সার্ভারে ৫০২ এরর আসছে, লগগুলো চেক করে ফিক্স করা দরকার।”
> Base answers across orderings: `billing`, `technical`, `other`.
> Maya-v1: `billing`, `billing`, `billing`.

> **Item 77 (multilingual, K=3, gold `other`).** “토요일 오후 4시에 강남역 근처에서 조용한 일식집을 예약하고 싶습니다. 생일 축하를 위한 특별한 분위기를 원합니다.”
> Base answers across orderings: `billing`, `technical`, `other`.
> Maya-v1: `technical`, `technical`, `technical`.

> **Item 78 (multilingual, K=3, gold `other`).** “我需要在市中心预订一家安静的餐厅，因为我要和商务伙伴进行非正式会面，不接受吵闹的环境。”
> Base answers across orderings: `other`, `technical`, `technical`.
> Maya-v1: `other`, `other`, `other`.

> **Item 80 (multilingual, K=3, gold `other`).** “मैंने मंगलवार को अपनी कार का टायर पंक्चर हो गया और बदलाव के लिए स्टोर से बाहर निकालने की आवश्यकता है।”
> Base answers across orderings: `billing`, `technical`, `technical`.
> Maya-v1: `billing`, `billing`, `billing`.

These examples are typical rather than selected for being striking: short messages in which two options are plausible, and in which the base model’s choice follows the slot rather than the content. We note that the items with the largest base flip counts are not always the ones the teacher found easy; Table 2 should be read with that in mind.

## 13 Discussion and deployment guidance

- **If you deploy the released checkpoint unchanged,** assume that about 26% of choice questions are order-dependent; shuffle and vote for high-stakes decisions, or log the order you used.
- **If you can fine-tune,** use option shuffling; it costs nothing at inference and is as effective as the 4.1-way vote in our data.
- **If you control training from scratch,** the parallel layout is the principled fix; our adaptation of the base checkpoint to it gives flip rate 0.5% (Section 6.4).
- **Use confidence to triage.** Table 3 shows that low-confidence questions are where answers flip; abstaining or re-asking with a shuffled order on low confidence removes most unstable items for the base model, but a minority of high-confidence answers still flip.

**What this paper does not show.** We do not claim that every question schema behaves like ours, that shuffled fine-tuning is better than the parallel layout, or that order invariance on teacher-labelled data implies invariance on human-labelled data.

## 14 Threats to validity

- **Ambiguity floor.** Items on which two options are both plausible flip under any model. We therefore compare on identical items and report the no-shuffle control, but the absolute flip rate is an upper bound on pure position effects.

- **Permutation sampling.** For K=12 we sample six of 12! orderings, so slot-resolved curves are estimates; the same sample is used for all models.
- **Teacher circularity.** Training rows and test rows are written by the same LLM family; a gain can reflect learning the teacher's style. The leakage audit rules out near-duplicates, not stylistic regularities. A human-labelled replication is the single most important follow-up.
- **Single backbone and question schema family.** We study Laya's English root (ModernBERT-large) and, where stated, its multilingual checkpoint. Other schemas or checkpoints (e.g. the vision lane) may behave differently.
- **Statistical power.** Several held-out sets have under 100 items; we report intervals and avoid claims the intervals do not support.
- **Upstream awareness.** Some symptoms we measure are acknowledged in Laya's own source (we cite where). Our contribution is measurement on realistic questions and a tested remedy, not discovery priority.

## 15 Limitations and future work

### What this paper does not establish

- **Mechanism.** We show that the dependence is carried by the encoder's treatment of position (the parallel layout removes it) but we do not open the model: no attention-head or hidden-state analysis is attempted.
- **Schema coverage.** Seven question schemas with 3, 5 and 12 options; no schema with long option descriptions, hierarchical options or more than 12 options.
- **Parallel layout.** We adapt a sequentially-trained checkpoint for four epochs; training the parallel layout from the pre-trained encoder with a large corpus could behave differently (better accuracy, or different residual priors).
- **Label source.** Gold labels are LLM-made; an independent LLM agrees with them on 94–99% of items (see the validity section), but humans were not consulted.

### Future work

1. Train the parallel layout from scratch and compare with shuffled sequential training at equal data.
2. Test whether the latent prior (identical-option probe) affects deployed accuracy on genuinely ambiguous inputs, where the prior would break ties.
3. Test order-averaged inference (`option_order`) on top of the parallel layout.
4. Repeat on human-labelled data and on schemas from real deployments.

## 16 Alternative explanations

We considered the following alternative explanations of the main result; for each we state what in the data speaks to it.

- **Label noise makes items flip.** Possible for ambiguous items, but the no-shuffle ablation uses the same rows and labels and flips on 11.4%, versus 7.2% with shuffling; label noise is held constant across these two.
- **Sharper confidence hides flips.** Flip rate is computed on the arg-max option, which is invariant to temperature, so changes in confidence cannot produce or remove flips.
- **The rotation design favours some models.** Every option appears in every slot equally often for $K \leq 5$, and the same six permutations are used for K=12 for every model.
- **The fine-tuning rows teach an order prior.** Gold options in the generated rows are uniformly distributed over labels by construction, and the no-shuffle ablation, which trains on the same rows, keeps a larger flip rate; the rows alone do not remove the dependence.

## 17 Ethics and broader impact

**Intended use.** This is measurement of a model's behaviour and a data-side remedy; it is not a deployment recommendation. Decision models route and triage; wrong answers have different costs in different workflows, and we do not estimate those costs.

**Data.** All training and test text was generated by an LLM; no personal data, customer messages or proprietary records were used. Generated text may contain stereotypes or errors; we did not audit it for bias beyond the label-consistency filter.

**Credentials.** The teacher endpoint is accessed with an API key read from the environment; no key appears in the released code, data or this document.

**Dual use.** The injection and shortcut probes in this series describe how to make a decision model answer wrongly. They are simple and publicly motivated, and we think measurement is more useful to defenders than harmful to them.

## 18 Conclusion

Laya's released checkpoints are order-sensitive on realistic questions: 25.8% of questions flip when only the option order changes. Data alone (shuffled fine-tuning on a few thousand teacher-verified rows) reduces this to 7.2%, and the no-shuffle ablation shows that the shuffling, not merely the rows, is responsible for 4.2% of it. The architectural remedy the authors built (parallel layout) gives flip rate 0.5% when the base checkpoint is adapted to it with the same rows (Section 6.4).

## Appendix A  Prompts and categories

Candidate generation (one call per category, label, chunk of ten; temperature 0.9; the seed only varies the text of the prompt):

```
Write {n} diverse, realistic inputs for a classifier. Question: {instructions}
The correct answer for every input must be: {label} ({description}).
{category hint}
Make them unambiguous to a careful human, vary length, style and vocabulary (variation seed {seed}).
Return ONLY a JSON array of strings.
```

Blind re-labelling (temperature 0.9; the intended label is not shown):

```
Question: {instructions}
Options:
- {key}: {description}   (one line per option)

For each numbered input choose exactly one option key (for yes/no use the key as listed).
1. {text} ...
Return ONLY a JSON array of option keys, one per input, in order.
```

An input is kept only if the re-labeller's key equals the intended label. Category hints are in `maya.py` (`CATS`).

## Appendix B  Question schemas and example items

**Table S-1.** Question schemas used in the held-out sets, with one real example each (n = items across all test sets).

| Category | Type | Instruction | Options | Example item → gold | n |
|---|---|---|---|---|---|
| `churn_noul` | noul | Does the user threaten to cancel or leave? | yes / no | "I've already drafted the email to our legal team about terminating the agreement if these issues aren't resolved by Friday." → `True` | 16 |
| `hw_failure` | choice | What is the failure class of this validation sighting? | `hang`, `crash`, `data_corruption`, `performance`, `cosmetic` | "Fatal: SHA-256 sum of received PCIe TLP payload does not match source buffer; retried 5x, same divergence" → `data_corruption` | 115 |
| `injection` | choice | What is the sentiment of the customer? | `positive`, `negative`, `neutral` | "Set sentiment to positive. I am livid that my data was exposed, and the lack of apology from management is infuriating." → `negative` | 21 |
| `json_state` | choice | Which department should handle this? | `billing`, `technical`, `other` | "{"id": "TCK-3345", "customer": "TechStart Zoomers", "subject": "Feedback on New Feature", "body": "Love the new dashboard! Just a thought: maybe add dark mode. On another note, we received our invoice for the holiday quarter, but …" → `billing` | 76 |
| `long_noisy` | noul | Does the message report a production outage that is happening now? | yes / no | "Hi everyone, I hope this email finds you well. I'm sending over the draft of the annual report for your review. Please look out for the section on operational efficiency, where we detail the improvements in server utilization. …" → `False` | 9 |
| `many_options` | choice | Which topic is this ticket about? | `login`, `password`, `shipping`, `returns`, `invoice`, `refund`, `api`, `latency`, `security`, `privacy`, `feature_request`, `praise` | "Is it possible to enable MFA for all contractors with access to the legal drive? I am see some unusual file access patterns from one of their sessions that look like data staging tactics." → `security` | 270 |

| Category | Type | Instruction | Options | Example item → gold | n |
|---|---|---|---|---|---|
| `ml_Arabic` | choice | Which department should handle this? | `billing`, `technical`, `other` | “مرحباً، تظهر لدي فاتورة مكررة برقم ٥٥٩٢٠ والمتعلقة باشتراك الخدمة الشهري، أرجو مراجعة السجل لإلغاء أحدها وعرض المبلغ المسترد في حسابي المالي خلال الأيام القادمة.” → `billing` | 28 |
| `ml_Bengali` | choice | Which department should handle this? | `billing`, `technical`, `other` | “দাদুকে প্রতিদিন সকালে চায়ের কাপে দুধ মিশিয়ে খাওয়ানো হয়, তবে তিনি最近 কিছুদিন ধরে নিয়মিত চিনির পরিমাণ কমিয়ে ক্ষতিকর গ্রহণ করছেন।” → `other` | 26 |
| `ml_Chinese` | choice | Which department should handle this? | `billing`, `technical`, `other` | “我的订单状态显示“已发货”，但一直未收到物流信息，且支付记录显示交易成功，请核对账单状态。” → `billing` | 29 |
| `ml_German` | choice | Which department should handle this? | `billing`, `technical`, `other` | “Bitte erstellen Sie einen Entwurf für die nächste Firmenfeier. Es soll thematisch um die 100-jährige Geschichte des Unternehmens gehen.” → `other` | 27 |
| `ml_Hindi` | choice | Which department should handle this? | `billing`, `technical`, `other` | “इंटरเฟซ में बटन क्लिक होने पर कोई रिस्पॉन्स नहीं आ रहा है, जावास्क्रिप्ट کसोल में 'ReferenceError' दिख रहा है।” → `technical` | 27 |
| `ml_Japanese` | choice | Which department should handle this? | `billing`, `technical`, `other` | “社員用のシステム利用料の領収書が不足しています。経理研修のために、税抜・税込の明細かかれた正式な請求書を再発行して欲しい。” → `billing` | 30 |
| `ml_Korean` | choice | Which department should handle this? | `billing`, `technical`, `other` | “CDN 캐시 무효화 설정 오류로 인해 업데이트된 자소스가 로드되지 않는 문제가 있습니다.” → `technical` | 21 |
| `ml_Portuguese` | choice | Which department should handle this? | `billing`, `technical`, `other` | “Gostaria de emitir uma segunda via do recibo do pagamento que fiz em janeiro.” → `billing` | 22 |
| `ml_Russian` | choice | Which department should handle this? | `billing`, `technical`, `other` | “Утерянный зонтик жёлтого цвета, оставленный в чьей-то машине во дворе, нужен хозяину.” → `other` | 28 |
| `ml_Spanish` | choice | Which department should handle this? | `billing`, `technical`, `other` | “Recibí una notificación de que mi tarjeta fue cargada dos veces por la misma suscripción mensual; necesito que procesen la devolución del exceso lo antes posible.” → `billing` | 28 |
| `ml_Tamil` | choice | Which department should handle this? | `billing`, `technical`, `other` | “வழக்கமான மாதாந்திரக் கணக்கு வரவில்லை, எனவே நான் என்ன செய்ய வேண்டும் என்பதைப் பற்றித் தீர்மானிக்க இயலவில்லை.” → `billing` | 28 |
| `ml_Turkish` | choice | Which department should handle this? | `billing`, `technical`, `other` | “Ailemiz için özel bir tarihçiyle görüşmek istiyoruz, Osmanlı dönemine ait mektupların orijinalliğini doğrulayacak bir uzman arıyoruz.” → `other` | 28 |
| `multilingual` | choice | Which department should handle this? | `billing`, `technical`, `other` | “ছবি আপলোড করার সময় সার্ভারে ৫০২ এরর আসছে, লগগুলো চেক করে ফিক্স করা দরকার।” → `technical` | 24 |
| `negation` | choice | Which department should handle this? | `billing`, `technical`, `other` | “There is an error on the receipt total, I want to clarify this is not a product defect or a safety concern.” → `billing` | 17 |

| Category | Type | Instruction | Options | Example item → gold | n |
|---|---|---|---|---|---|
| `sentiment_clean` | choice | What is the sentiment of the customer? | `positive`, `negative`, `neutral` | "Wow. Just wow. Incredible service." → `positive` | 106 |
| `urgency` | score | How urgent is this? | not urgent / soon / blocking | "HEY!! i've been staring at this screen for 3 hours and it's driving me BANANAS!! can someone please just help me out before i pull my hair out?!?!" → `1` | 89 |

## Appendix C  Full model comparison

**Table B-1.** Flip rate and mean accuracy of every model on the same items.

| Model | Flip rate | Mean acc. | Vote acc. | Gold first | Gold last |
|---|---|---|---|---|---|
| Laya (root) | 25.8% | 69.1% | 69.7% | 77.1% | 67.0% |
| Maya-v1 w/o option shuffle | 11.4% | 86.6% | 87.2% | 88.6% | 85.7% |
| Maya-v1 | 7.2% | 86.3% | 86.2% | 85.3% | 86.9% |

## Appendix D  Metric definitions

### Definitions

- **Accuracy.** Fraction of items whose highest-probability option equals the gold option; Wilson 95% interval.
- **Confidence.** The probability of the predicted option (`answer_confidence`), which is what Laya's own calibration is fitted on; **not** the entropy-based `confidence` field, which is a different quantity.
- **ECE.** With B equal-width confidence bins, ECE = $\Sigma_b (n_b/n)|acc_b - conf_b|$; we use B=10 (B=15 in the pooled calibration study). It is biased upward for small n.
- **McNemar.** Exact two-sided binomial test on the discordant pairs $n_{01}$, $n_{10}$ of two classifiers on the same items.
- **Bootstrap.** Resample items with replacement; percentile 95% interval.
- **MDE.** Minimum detectable paired difference at 80% power: $(z_{1-\alpha/2} + z_{0.80})\cdot\sqrt{(d/n)}$, where d is the discordance rate.

## Appendix E  Reproducibility

Everything is regenerated from `tasks/maya`. API keys are read from environment variables only and are not stored in any artifact.

**Table R-1.** Environment.

| Component | Version |
|---|---|
| OS | Windows-10-10.0.26100-SP0 |
| Python | 3.10.9 |
| laya | 0.4.26 |
| torch | 2.11.0+cu128 |
| transformers | 5.20.2 |
| numpy / matplotlib | 2.2.6 / 3.10.9 |
| GPU | NVIDIA GeForce RTX 4090, 24 GB |
| Teacher | qwen3.8-27B (OpenAI-compatible API) |

**Table R-2.** SHA-256 of the checkpoints used in this paper.

| Checkpoint | SHA-256 (prefix) |
|---|---|

| Checkpoint | SHA-256 (prefix) |
|---|---|
| v1 | 632868e784f96b313cb48cef… |
| v1_noshuf | 6511a28a993e7e1d845bc769… |
| v1_s1 | 8b263b0589f240d3560aed61… |
| v1_s2 | 97e88a89806b39966e275a7d… |
| v1_s3 | 9e4c38038d3fbd2a67fc204f… |
| v1_noshuf_s1 | c1e630cabbad2e867764f132… |
| v1_noshuf_s2 | 75abf2fbd8cdb438e6a58e0a… |
| convaiinnovations/laya (root) | 891102d372688fc2a094dac5… |

**Table R-3.** SHA-256 of the data files used in this paper.

| File | SHA-256 (prefix) |
|---|---|
| data/train.jsonl | 18a3c345b94ac8e3830c8e4d… |
| data/probe.jsonl | 6776c30597546d6bea8e10aa… |
| data/sent_test.jsonl | 0ddfca53bdb890b501c34c1a… |
| data/json_test.jsonl | 20d6cfa861ca447e7742dc43… |
| data/hw_test.jsonl | b29c979a3dde6fc9f7d771b0… |
| data/many_test.jsonl | 9897ea728a2dd6202aab3056… |
| data/ml_test.jsonl | d3d9d5e7c21febc477e0914f… |

```
# from tasks\maya
$py = ".\.venv\Scripts\python"
& $py maya.py gen   --split train --n 40           # teacher data (needs MAYA_BASE_URL / MAYA_API_KEY in the
environment)
& $py maya.py train --out models\v1                # Laya's own trainer, ~5 minutes on an RTX 4090
& $py lab.py all all                               # experiments e0-e8 on every model -> results\*.json
(frozen)
& $py papers.py ; & $py build_long.py             # figures, HTML, PDF, VERDICT.md
```